\documentclass[11pt,a4paper]{article}
\usepackage[hyperref]{acl2020}
\usepackage{times}
\usepackage{latexsym}

\usepackage{microtype}

\usepackage{tabularx}
\usepackage{multicol}
\usepackage{multirow}

\usepackage{color, soul}

\usepackage{graphicx}
\usepackage{subfig}

\usepackage{amssymb}
\usepackage{pifont}
\aclfinalcopy 

\title{Fine-grained Emotion and Intent Learning in Movie Dialogues}

\author{Anuradha Welivita, Yubo Xie and Pearl Pu\\
  School of Computer and Communication Sciences \\
  École polytechnique fédérale de Lausanne \\
  Switzerland \\
  {\tt \{kalpani.welivita,yubo.xie,pearl.pu\}@epfl.ch} 
  \\}

\date{}

\begin{document}
\maketitle
\begin{abstract}
We propose a novel large-scale emotional dialogue dataset, consisting of 1M dialogues retrieved from the OpenSubtitles corpus and annotated with 32 emotions and 9 empathetic response intents using a BERT-based fine-grained dialogue emotion classifier. This work explains the complex pipeline used to preprocess movie subtitles and select good movie dialogues to annotate. We also describe the semi-supervised learning process followed to train a fine-grained emotion classifier to annotate these dialogues. Despite the large set of labels, our dialogue emotion classifier achieved an accuracy of $65\%$ and was used to annotate 1M emotional movie dialogues from OpenSubtitles. This scale of emotional dialogue classification has never been attempted before, both in terms of dataset size and fine-grained emotion and intent categories. Visualization techniques used to analyze the quality of the resultant dataset suggest that it conforms to the patterns of human social interaction.

\end{abstract}

\section{Introduction}
\label{sec:introduction}

Understanding emotion in human social conversations or chitchat has gained popularity in the natural language processing community due to its usefulness in developing human-like conversational agents. Emotions revealed in social chitchat are rather complex. It has many categories of emotions to distinguish due to subtle variations present in human emotion. For example, \textit{Sadness} and \textit{Disappointment} are pursued and dealt differently in human conversations. Also, the listeners' reaction to emotions is not always a straightforward mirroring effect of the speakers' emotions. Rather it can be more neutral and convey a specific intent, as is evident from the dialogue example in Table \ref{table:example}. 

\begin{table}[ht!]
\centering
\begin{tabularx}{\linewidth}{|r X|}
\hline
Speaker:&\textit{I’ve been hearing some strange noises around the house at night.} \textbf{(Afraid)}\vspace{0.2em}\\
Listener:&\textit{oh no! That’s scary! What do you think it is?} \textbf{(Neutral: Acknowledging; Questioning)}\vspace{0.2em}\\
Speaker:&\textit{I don’t know, that’s what’s making me anxious.} \textbf{(Anxious)}\vspace{0.2em}\\
Listener:&\textit{I’m sorry to hear that.} \textbf{(Neutral: Sympathizing)}\vspace{0.2em}\\
\hline
\end{tabularx}
\caption{An example dialogue showing that the listener's reactions to emotions do not always mirror the speaker's emotions.}
\label{table:example}
\end{table}

\begin{figure*}[ht!]
\centering
\includegraphics[width=\linewidth]{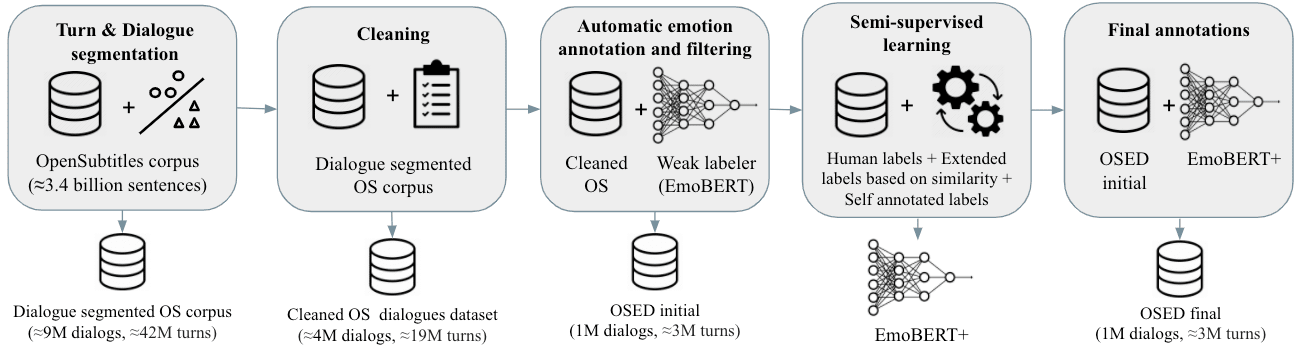} 
\caption{Steps for curating the OSED dataset.}
\label{fig:steps}
\end{figure*}


Welivita and Pu \shortcite{taxonomy} have analyzed listener responses in the EmpatheticDialogues dataset \cite{empatheticdialogues} and discovered 9 listener specific empathetic response intents contained in emotional dialogues: \textit{Questioning}; \textit{Agreeing}; \textit{Acknowledging}; \textit{Sympathizing}; \textit{Encouraging}; \textit{Consoling}: \textit{Suggesting}; \textit{Wishing}; and \textit{Neutral} (include other intents with neutral emotion such as \textit{expressing an opinion}, \textit{advising}, and \textit{disagreeing}). They have automatically annotated the EmpatheticDialogues dataset \cite{empatheticdialogues} with 32 fine-grained emotions and the 9 empathetic response intents and discovered frequent emotion-intent exchange patterns in human social conversations. They observe that this type of dataset tagged with fine-grained emotions and response intents could train neural chatbots to generate empathetically appropriate responses conditioned on a selected emotion or intent. However, for this purpose, a large-scale emotion and intent labeled dataset is even more desirable. Curating such a dataset is technically challenging because 1) annotating such a large-scale dataset require human labor that is costly, and 2) given the fine-granularity of the emotion and intent labels, the human labeling task is more difficult compared to more generic \textit{Angry}-\textit{Happy}-\textit{Sad}. As a result, existing manually labeled emotional dialogue datasets such as IEMOCAP \cite{iemocap}, MELD \cite{meld}, and DailyDialogue \cite{dailydialogue} are smaller in scale and contain only a limited set of emotions (emotions derived from basic emotion models such as the Ekman's), with simpler dialogue responding strategies, or both. Also, existing datasets often contain a label \textit{Neutral} or \textit{Other} for responses that do not convey emotion, which introduces vagueness and limits the ability of automatic agents that use such datasets in learning useful response strategies. 


To fill the above gap, we curate a novel large-scale dialogue dataset, OSED (OpenSubtitles Emotional Dialogues), containing 1M emotional dialogues from movie subtitles, in which each dialogue turn is automatically annotated with 32 fine-grained emotions and 9 empathetic response intents. Movie subtitles well approximate human social conversations and how emotion is handled in them. It is one of the major sources to learn emotional variations and corresponding response strategies. To reduce the cost of human labeling and the complexity of labeling dialogues with fine-grained emotions and intents, we devise a semi-automated human computation task to collect fine-grained emotion and intent labels for a small set of movie dialogues (9K). We then follow a semi-supervised approach to expand the labeled data and train a dialogue emotion classifier to automatically annotate 1M emotional dialogues. 

The process of curating the dataset consists of several stages. First, we apply automatic turn and dialogue segmentation methods on movie subtitles in the OpenSubtitles (OS) corpus \cite{opensubtitles} and obtain close to 9M dialogues. After data cleaning and removing duplicates, we reduce its size to 4M. Then, we apply a weak labeler, EmoBERT (a BERT-based sentence-level classifier) trained on the EmpatheticDialogues dataset \cite{empatheticdialogues}, to label utterances in OS dialogues and filter 1M emotional dialogues (OSED initial). Thirdly, with semi-supervised learning methods, we refine EmoBert and obtain EmoBert+, a more advanced dialogue emotion classifier trained on OS dialogues. To evaluate EmoBert+, we compare it with FastText. The former is more accurate than FastText. Finally, we use EmoBert+ to label dialogues in OSED initial to obtain the final 1M OSED dataset. We evaluate the quality of the resultant dataset by visually inspecting the emotion-intent flow patterns that occur in the dataset and checking if they conform to the patterns of human social conversations discovered in existing work \cite{see,taxonomy}. Figure \ref{fig:steps} summarizes the process of creating OSED. The data curation pipeline we follow substantially reduces the cost of human labor, while ensuring quality annotations. 

Our contributions in this paper are three-fold. 1) We curate a dialogue dataset, OSED, containing 1M emotional dialogues labeled with 32 fine-grained emotions and 9 empathetic response intents. Compared to existing dialogue datasets tagged with emotions, OSED is more general-purpose, significantly larger, and contains more fine-grained emotions and empathetic response strategies. 2) We outline the complex pipeline used to derive this dataset and evaluate the annotation quality using visualization methods. 3) We release our fine-grained emotion classifier used to annotate the OSED dataset, which can be used as a general-purpose classifier capable of recognizing fine-grained emotions and empathetic response intents in social chitchat.

\section{Literature Review}
\label{sec:lit_review}

\begin{table*}[ht!]
\centering
\begin{tabularx}{\textwidth}{|p{2.8cm}|p{6.5cm}|r|r|c|}
\hline
Dataset & Labels & No. of & No. of  & Publicly \\
& & dialogues & utterances & available\\
\hline
IEMOCAP \cite{iemocap} \vspace{2mm}& \textit{Joy}, \textit{Sadness}, \textit{Anger}, \textit{Frustrated}, \textit{Excited}, and \textit{Neutral} & $151$ & $7,433$ & 
\ding{51}\\
MELD \cite{meld} \vspace{2mm}& \textit{Joy}, \textit{Surprise}, \textit{Sadness}, \textit{Anger}, \textit{Disgust}, \textit{Fear}, and \textit{Neutral} & $1,433$ & $13,708$ & \ding{51}\\
DailyDialogue \cite{dailydialogue} \vspace{2mm}& \textit{Joy}, \textit{Surprise}, \textit{Sadness}, \textit{Anger}, \textit{Disgust}, \textit{Fear}, and \textit{Neutral} & $12,218$ & $103,607$ & \ding{51}\\
EmotionLines \cite{emotionlines} \vspace{2mm}& \textit{Joy}, \textit{Surprise}, \textit{Sadness}, \textit{Anger}, \textit{Disgust}, \textit{Fear}, and \textit{Neutral} & $1,000$ & $14,503$ & \ding{51}\\
EmoContext \cite{emocontextdataset}\vspace{2mm}& \textit{Joy}, \textit{Sadness}, \textit{Anger}, and \textit{Other} & $38,421$ & $115,263$ & \ding{51}\\
Twitter customer support \cite{CustomerSupport}& Customer emotions: \textit{Confusion}; \textit{Frustration}; \textit{Anger}; \textit{Sadness}; \textit{Happiness}; \textit{Hopefulness}; \textit{Disappointment}; \textit{Gratitude}; \textit{Politeness}; and Agent emotional techniques: \textit{Empathy}; \textit{Gratitude}; \textit{Apology}; \textit{Cheerfulness}\vspace{2mm}& $2,413$ & $\approx14,078$ & 
\ding{55}\\
Empathetic Dialogues \cite{empatheticdialogues,taxonomy}\vspace{2mm}& 32 fine-grained emotions and 9 empathetic response intents: \textit{Questioning}; \textit{Agreeing}; \textit{Acknowledging}; \textit{Sympathizing}; \textit{Encouraging}; \textit{Consoling}; \textit{Suggesting}; \textit{Wishing}, \textit{Neutral}. & $24,850$ & $107,220$ & 
\ding{51}\\

OSED & 32 fine-grained emotions and 9 empathetic response intents. & $1M$ & $3,488,300$ & 
\ding{51}\\
\hline
\end{tabularx}
\caption{Comparison of emotion annotated dialogue datasets available in the literature against OSED.}
\label{table:datasets}
\end{table*}



IEMOCAP \cite{iemocap}, MELD \cite{meld}, DailyDialogue \cite{dailydialogue}, EmotionLines \cite{emotionlines}, and EmoContext \cite{emocontextdataset} are some existing dialogue datasets with emotion labels. Table \ref{table:datasets} shows a summary of the size and the labels present in these datasets. Even though these datasets approximate social chitchat quite well, they are limited in size and are labeled with only a small set of emotions without any response strategies. Several state-of-the-art dialogue emotion classifiers such as CMN \cite{cmn}, ICON \cite{icon}, IAAN \cite{iaan}, DialogueRNN \cite{dialoguernn}, and DialogueGCN \cite{dialoguegcn} were proposed based on the above datasets. They report accuracy in the range of $56-65\%$ on a limited set of emotion labels. Though these dialogue emotion classifiers report high accuracy, they are incapable of identifying fine-grained emotions and specific response strategies in human conversations, which makes us unable to use them for our purpose of curating an automatically labeled dialogue corpus. 



Herzig et al. \shortcite{CustomerSupport} detected both customer emotions and agent emotional techniques (e.g., \textit{Apology}, \textit{Empathy}) in customer support dialogues. They curated a dialogue dataset from two customer support Twitter accounts and manually annotated the customer turns with one of 9 emotions and the agent turns with one of 4 emotional techniques. But emotions expressed by customers in social media service dialogues are mainly negative (e.g. \textit{anger}, \textit{frustration}). Customer service agents are also instructed to behave in a certain way, which is different from daily chitchat. Social chitchat can include a variety of positive and negative emotions and a variety of different response intents.


The EmpatheticDialogues dataset \cite{empatheticdialogues} contains 25K open-domain dialogues grounded on 32 emotions. The 32 emotions range from basic emotions derived from biological responses \cite{ekman,plutchik} to larger sets of subtle emotions derived from contextual situations \cite{skerry}. The dialogues were collected using Amazon Mechanical Turk, instructing the speakers to start a dialogue picking one of 32 emotions and the listeners to respond empathetically. A major limitation in this dataset is that individual dialogue turns are not labeled, and it does not identify specific intents that listeners use when responding to speakers' emotions. Welvita and Pu \shortcite{taxonomy} manually analyzed a subset of the listener turns in EmpatheticDialogues and identified 9 listener specific response intents. They developed a sentence level weak labeler using which they annotated the entire dataset with 32 emotions and 9 empathetic response intents. However, due to the limited size of EmpatheticDialogues, it is difficult to use this for data-intensive applications.

To address the above limitations, we curate OSED containing 1M movie dialogues that approximate social emotional conversation. We label each dialogue turn with 32 emotions and 9 empathetic response intents using our own dialogue emotion classifier. Table \ref{table:datasets} compares OSED to existing emotion annotated dialogue datasets in the literature. Our dialogue emotion classifier trained on movie subtitles achieves state-of-the-art performance despite the large set of labels.

\section{Methodology}
\label{sec:methodology}

This section describes the dialogue selection process, design of the human annotation task, sentence embeddings techniques used to expand human-labeled dialogues, the architecture, and the semi-supervised training process of the classifier used to produce the resultant 1M annotated OSED dataset. 

\subsection{Dialogue Curation from Movie Subtitles}

The OpenSubtitles 2018 corpus consists of 3.7M movie and TV subtitles. It is comprised of 3.4B sentences and 22.2B tokens. It is an excellent source to learn emotional variations in dialogue and corresponding response mechanisms since movie and TV dialogues well approximate social chitchat and how humans handle emotional events. However, movie subtitles do not contain an explicit dialogue turn structure and specific indicators where one dialogue ends and the next dialogue begins. To overcome the first issue, we reproduced the work by Lison and Meena \shortcite{turn_segmentation} to build an SVM based classifier that determines if two consecutive sentences are part of the same dialogue turn. Our classifier achieved a segmentation accuracy of $76.69\%$, which is close to the accuracy of $78\%$ that the authors claim. The set of features that gave the best turn segmentation accuracy are: 1) unigram and bi-gram features of adjacent sentences after lemmatization; 2) first and final tokens of adjacent sentences; 3) first and final bi-grams of adjacent sentences; 4) whether the 2 sentences belong to the same subtitle block or not (boolean); 5) genre of the movie (\textit{Drama}, \textit{Crime}, \textit{Musical} etc.); 6) sentence density of the subtitles file (no. of sentences / subtitle duration); and 7) quadratic combinations of the above features with itself and the rest.

After performing turn segmentation on the OpenSubtitles corpus, we divided the turns into separate dialogues based on a simple heuristic. If the difference between the end time of the previous turn and the start time of the current turn is more than 5 seconds, we take these two turns as belonging to 2 different dialogues. An exception occurs if this timestamp information is missing in at least one of the turns. In this case, we assume that these two turns appear in the same subtitle block and consider them as belonging to the same dialogue. This way, we formed 9M dialogues from the OpenSubtitles corpus altogether. The choice of 5 sec.s to separate dialogues is explained in Appendix \ref{sec:app_params}. 

To further clean the dialogues, we removed character names, repetitive dialogue turns, turns that start with ``previous on..." (monologue at the beginning of TV episodes), turns with character length less than $2$ or greater than $100$, turns with an alphabetic proportion less than $60\%$, and turns with a lot of repetitive tokens. When a dialogue turn was removed, all the turns following that turn were also removed from the dialogue to maintain consistency. After that, all the dialogues left with only one turn were removed from the corpus. This resulted in a cleaned OS dialogue dataset consisting of 4M dialogues. This dataset is mainly used to train our fine-grained dialogue emotion classifier.



To filter the most emotional dialogues from the cleaned OS dialogues dataset, we employed a weak labeler, EmoBERT (a BERT transformer-based sentence level classifier) trained on 25K situation descriptions from EmpatheticDialogues \cite{empatheticdialogues} tagged with 32 emotion classes, and 7K listener utterances tagged with 9 empathetic response intents \cite{taxonomy}. The classifier had a high top-1 classification accuracy of $65.88\%$. We call it a weak labeler since it predicts emotion or intent only at sentence-level and is trained on a different dataset other than OS. We filtered the top 1M dialogues having the highest label confidence as predicted by this classifier to form the 1M OSED (initial) dataset. The statistics of the OSED dataset is given in Table \ref{table:osed}. 

\begin{table}[ht!]
\centering
\begin{tabularx}{\linewidth}{|X|r|}
\hline
Total no. of dialogues & $1,000,000$ \\
Total no. of turns & $2,829,426$ \\
Total no. of tokens & $39,469,825$ \\
Avg. no. of turns per dialogue & $2.83$ \\
Avg. no. of tokens per dialogue & $39.47$ \\
Avg. no. of tokens per turn & $13.95$ \\

\hline
\end{tabularx}
\caption{Statistics of the OSED dataset.}
\label{table:osed}
\vspace{-5pt}
\end{table}

\subsection{Human Computation}

To train a dialogue emotion classifier that can identify both fine-grained emotions and empathetic response intents, we devised an Amazon Mechanical Turk (AMT) experiment to collect an initial set of ground truth labels for OS dialogues. But annotating dialogue turns with one of $41$ labels is a daunting task. To make the task less exhaustive, we devised a semi-automated approach using our weak labeler, EmoBERT. By applying EmoBERT on each turn of the cleaned OS dialogue dataset, we filtered the turns having prediction confidence $\geq 0.9$, along with their dialogue history. Next, we ranked these dialogues according to their readability and selected the highest readable dialogues from each class to be labeled. This is to reduce the time spent by the workers in having to read long and complicated dialogues. The steps followed in computing dialogues' readability are included in Appendix \ref{sec:app_readability}. Workers had to select a label from the top-3 predictions made by EmoBERT. If none of the top-3 predictions matched, they could manually specify the correct class. If the correct emotion or intent was not present among the $41$ labels we provide, they could manually enter any new emotion or intent they think is appropriate. The AMT task's user interface design is included in Appendix \ref{sec:app_interface}.

After ranking the dialogues according to readability, we selected the top 250 dialogues in each category for the AMT task. We bundled 15 dialogues in a HIT with 5 quiz questions that served as checkpoints to evaluate the crowd workers' quality. Situation descriptions from the EmpatheticDialogues dataset for which we already knew the emotion labels were used to formulate the quiz questions. Finally, we obtained dialogues where we had 2 out of 3 worker agreement, which resulted in $8,913$ dialogues altogether. Table \ref{table:mturk_stats} shows the results of the AMT task. 


\begin{table}[ht!]
\centering
\begin{tabularx}{\linewidth}{|X|p{2.5cm}|}
\hline
Total no. of dialogues & $10,250$\\
No. of dialogues labeled & \\
with majority vote &$8,913 (86.96\%)$\\
Inter-annotator agreement (Fleiss' Kappa) & $0.46$ (moderate agreement)\\ 
Avg. time taken per HIT & $20.74$ min.\\
\% of times workers got 3/5 & \\
quiz questions correct & $77.75\%$\\
\hline
\end{tabularx}
\caption{AMT task results.}
\label{table:mturk_stats}
\end{table}

By allowing crowd workers to specify any new emotion or intent class, we were able to gather suggestions on new intent classes such as \textit{Greeting}, \textit{Declaration}, \textit{Commanding}, \textit{Hesistant}, and \textit{Accusing} that were not already present among the labels provided. They are listed with corresponding examples in Appendix \ref{sec:app_newlabels}. Since incorporating these new intents requires thorough validation and a sufficient number of examples from the entire dataset to be grounded as frequently appearing intents in social chitchat, we leave it as future work.

\subsection{Distant Learning Using Dialogue Embeddings}

To extend the training data obtained from the AMT task, we used the Sentence-BERT (SBERT) approach proposed by Reimers and Gurevych \shortcite{siamese}, which uses siamese and triplet network structures to derive semantically meaningful sentence embeddings that can be compared using cosine-similarity. Using this approach, we obtained semantically similar dialogues to those annotated by crowd workers and tagged them with the same class label. Among several models the authors have proposed, we used the \textit{roberta-base-nli-stsb-mean-tokens} model, fine-tuned on the NLI \cite{A_large_annotated_corpus_for_learning_natural_language_inference} and STS benchmark (STSb) \cite{SemEval-2017_Task1_Semantic_Textual_Similarity_Multilingual_and_Crosslingual_Focused_Evaluation} datasets, since it has a high score in the STS benchmark, and is more efficient to use than \textit{roberta-large}. 

Before proceeding, we left $20\%$ of the crowd-annotated dialogues, balanced across all class labels, as testing data to test our classifier's performance over different iterations. Then, we followed the following steps to extend the rest of the dialogues using SBERT. 1) Using the SBERT model, first, we computed dialogue turn embeddings (each with a vector representation of 768 dimensionalities) for all the turns ($\approx$19M) in the cleaned OS dataset. 2) Then, we calculated dialogue embeddings for both human-annotated and unlabeled dialogues from the cleaned OS dialogues dataset. For this, we applied a decaying weight starting from the last turn and took the weighted average of the turn embeddings of each dialog. We used half decaying, i.e, if we have a dialogue with turn embeddings $v_1$, $v_2$, and $v_3$, the final dialogue embedding would be $(4/7)v_3 + (2/7)v_2 + (1/7)v_1$. 3) Next, we calculated the cosine-similarity between annotated and unlabeled dialogue embeddings and ranked the results. 4) Finally, we applied a similarity threshold and obtained all the unlabeled dialogues with a cosine similarity that exceeds this threshold and tagged them with the same crowd annotated class label. Here, we used a threshold of $0.92$.

    
    

We could extend the original crowd annotated dialogue dataset by $3,196$ more dialogues with distantly annotated class labels using the above method. The selection of parameters, such as the decaying weights, is explained in Appendix \ref{sec:app_params}.

\subsection{The Fine-grained Emotion Classifier}

\begin{figure}[ht!]
\centering
\includegraphics[width=\linewidth]{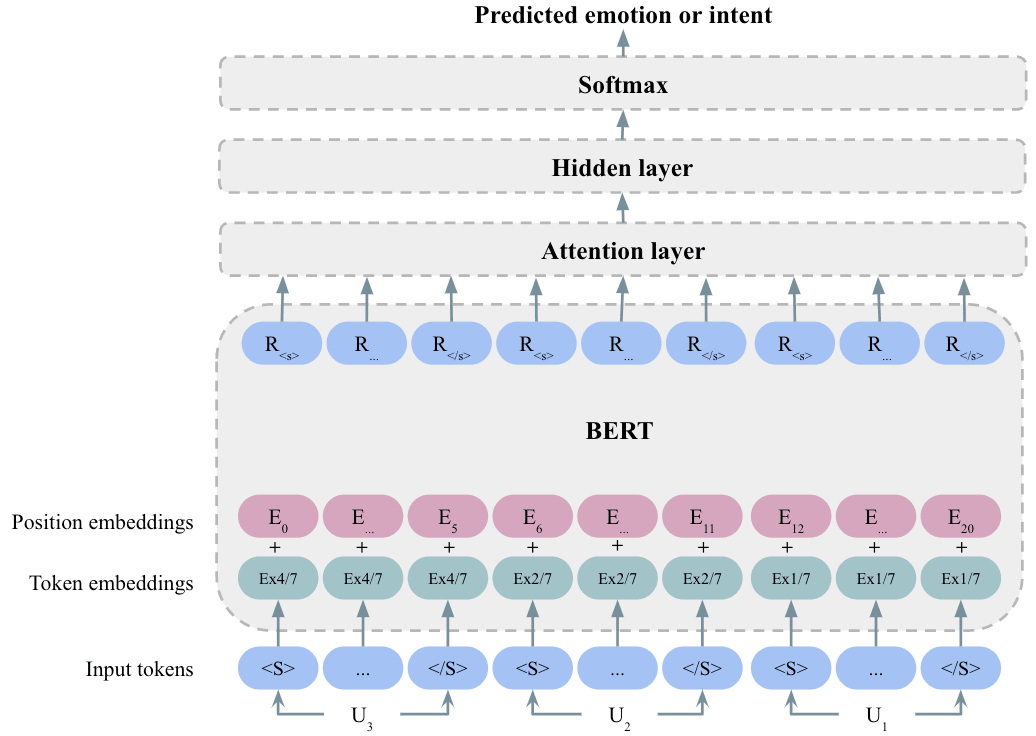}
\caption{EmoBERT+ architecture.}
\label{fig:eb+}
\end{figure}

We use the curated data to train a fine-grained dialogue emotion classifier, EmoBERT+ (EB+), to annotate the movie dialogues. EB+ consists of a representation network that uses the BERT architecture, an attention layer that aggregates all hidden states at each time step, one hidden layer, and one softmax layer. We use the BERT-base architecture with 12 layers, 768 dimensions, 12 heads, and 110M parameters as the representation network. We initialize it with weights from the pre-trained language model RoBERTa \cite{roberta}, which is proven to have superior performance than pre-trained BERT. We feed in a dialogue turn along with the preceding context in the reverse order as input to the representation network. To give more importance to the dialogue turn for which prediction has to be made and the turns that immediately precede it, we multiply the token embeddings belonging to each turn by a decreasing weight factor. As depicted in Figure \ref{fig:eb+}, for a given token, its input representation is constructed by summing the corresponding token embedding multiplied by the weighting factor and its position embedding. 

\begin{table*}[ht!]
\centering
\begin{tabularx}{\textwidth}{|p{2.75cm}|X|r|r|r|r|}
\hline
Model & Training data & Precis. & Recall & F1 & Acc.\\
\hline

FastText (base) & dialogues from AMT (5K) & $00.78$ & $04.73$ & $01.32$ & $06.36$\\

FastText (1\textsuperscript{st} iter.) & 5K + Similar dialogues from SBERT (3K) & $07.18 $ & $10.28$ & $05.65$ & $11.20$\\

FastText (2\textsuperscript{nd} iter.) & 5K + 3K + Self-labeled dialogues (4K) & $09.76$ & $11.19$ & $07.37$ & $12.04$\\

FastText (2\textsuperscript{nd} iter., extended) & 5K + 3K + 4K + Similar self-labeled dialogues from SBERT (2K) & $\textbf{13.23}$ & $\textbf{15.06}$ & $\textbf{10.89}$ & $\textbf{15.53}$\\

\hline

EB+ (base) & dialogues from AMT (5K) & $63.33$ &	$63.94$ &	$63.28$ &	$\textbf{65.17}$\\

EB+ (1\textsuperscript{st} iter.) & 5K + Similar dialogues from SBERT (3K) & $63.55$ & $63.54$ & $62.92$ & $64.55$\\

EB+ (2\textsuperscript{nd} iter.) & 5K + 3K + Self-labeled dialogues (4K) & $63.96$ & $64.27$ & $63.61$ & $64.88$\\

EB+ (2\textsuperscript{nd} iter., extended) & 5K + 3K + 4K + Similar self-labeled dialogues from SBERT (2K) & $\textbf{64.11}$ & $\textbf{64.59}$ & $\textbf{63.86}$ & $65.00$\\

\hline
\end{tabularx}
\caption{Performance results of the FastText (baseline) and EB+ classifiers over the semi-supervised learning iterations. All scores are reported on the OS test dataset. F1-score reported here is the macro-F1 score.}
\label{table:results}
\end{table*}

We split the original crowd annotated dialogues to $60\%$ training, $20\%$ validation, and $20\%$ testing, i.e., $5,390$ dialogues for training, $1,785$ for validation, and $1,777$ for testing and trained several models of EB+ iteratively following a semi-supervised learning approach. First, \textit{EB+ base} was trained only on the crowd annotated dialogues from AMT. Next, we trained a second model, \textit{EB+, 1\textsuperscript{st} iteration}, by combining the crowd annotated dialogues with the similar dialogues found using SBERT. 

After the 1\textsuperscript{st} iteration, we applied the classifier on unlabeled turns in cleaned OS dialogues and obtained the top-$100$ high confidence turns in each category along with their preceding context. It resulted in $4,100$ self-labeled dialogues, which were incorporated into the training dataset. Next, we trained \textit{EB+, 2\textsuperscript{nd} iteration} with the training dataset extended with the self-labeled dialogues. Finally, another model \textit{EB+, 2\textsuperscript{nd} iteration extended} was trained by re-applying the SBERT approach to obtain similar dialogues to the self-labeled dialogues that were newly incorporated in the 2\textsuperscript{nd} iteration. In this stage, we could incorporate $2,118$ additional dialogues having cosine-similarity above a threshold of $0.9$. We trained each model using a batch size of $256$ for $10$ epochs and obtained the one giving the lowest classification loss in the validation set. We used $100$ as the maximum length of input tokens and a learning rate of $2\times10^{-5}$. 


To compare EB+'s performance against a standard baseline, we followed the same semi-supervised learning steps to train a FastText classifier with default configuration \cite{fasttext}. In this, the dialogue turn for which the prediction has to be made was concatenated with preceding dialogue turns and fed as one input sequence to the FastText classifier. Finally, this experiment's best performing model was used to annotate all the turns in the OSED dataset automatically.

\section{Results}

The EB+ classifier's performance results over different iterations on the OS test dataset having $1,777$ dialogues distributed equally across classes are denoted in Table \ref{table:results}. The final model \textit{EB+, 2nd iteration extended}, trained on 14K labeled dialogue turns altogether, performs the best as denoted by precision, recall, and macro-F1 metrics. Even the accuracy of the final model does not deviate significantly from previous iterations. It could be noted that compared to FastText, the semi-supervised learning iterations could only yield minor improvement in EB+ since it already performs significantly well when trained only on the 5K crowd-annotated dialogue turns. The performance improvement in FastText is significant over the semi-supervised learning iterations; however, it is still notably low compared to EB+. Performance comparison of the best performing EB+ model with the state-of-the-art dialogue emotion classifiers is included in Appendix \ref{sec:app_performance_comparison}. A direct comparison with the state-of-the-art cannot be made since existing classifiers can only predict a limited number of emotion labels.

\section{OSED Quality Analysis}
\label{sec:quality}

\begin{figure*}%
    \centering
    
    \subfloat[\centering Beginning emotion: Joyful]{{\includegraphics[width=0.45\textwidth]{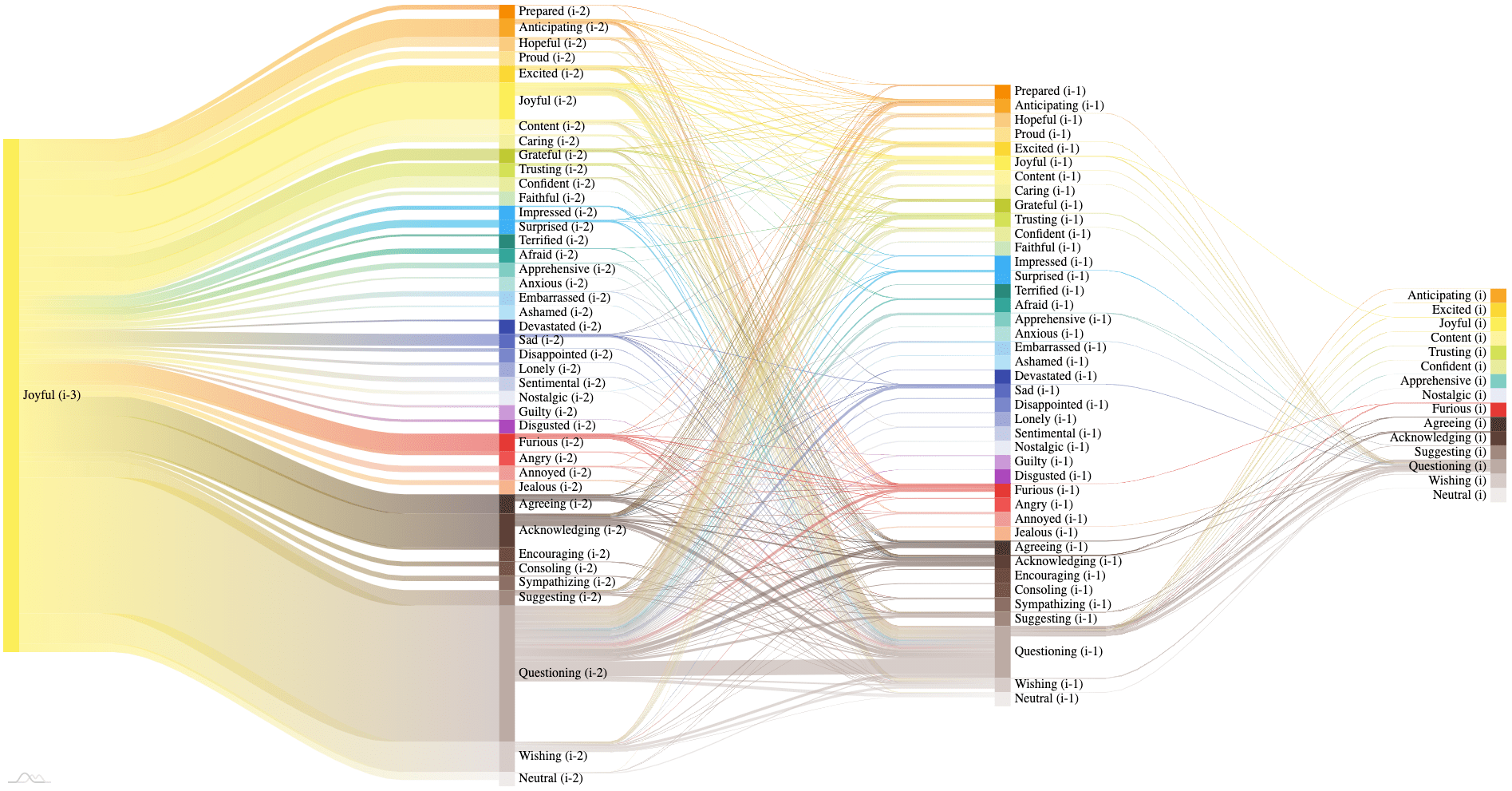} }}%
    \qquad
    \subfloat[\centering Beginning emotion: Surprised]{{\includegraphics[width=0.45\textwidth]{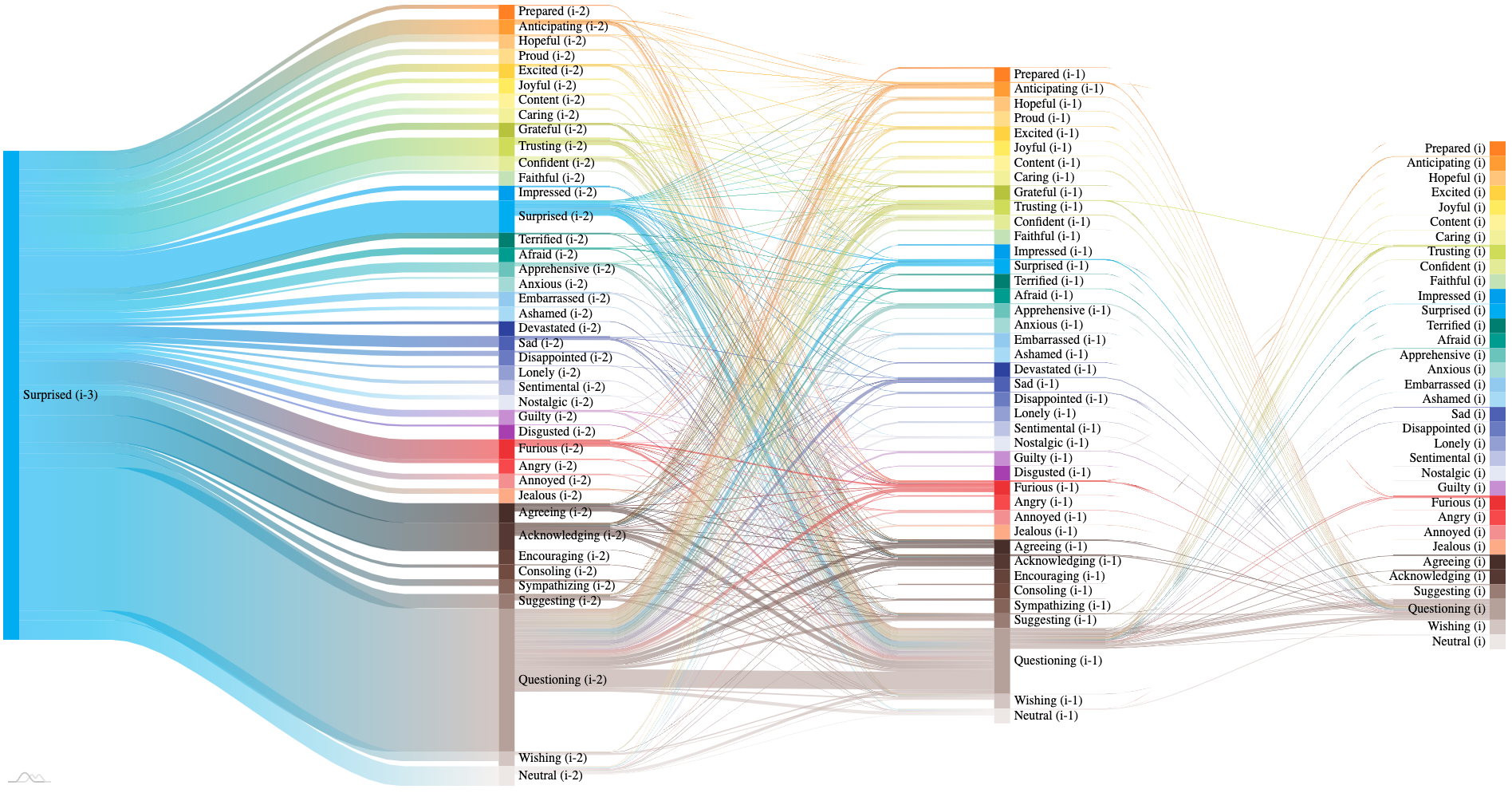} }}%
    
    \subfloat[\centering Beginning emotion: Sad]{{\includegraphics[width=0.45\textwidth]{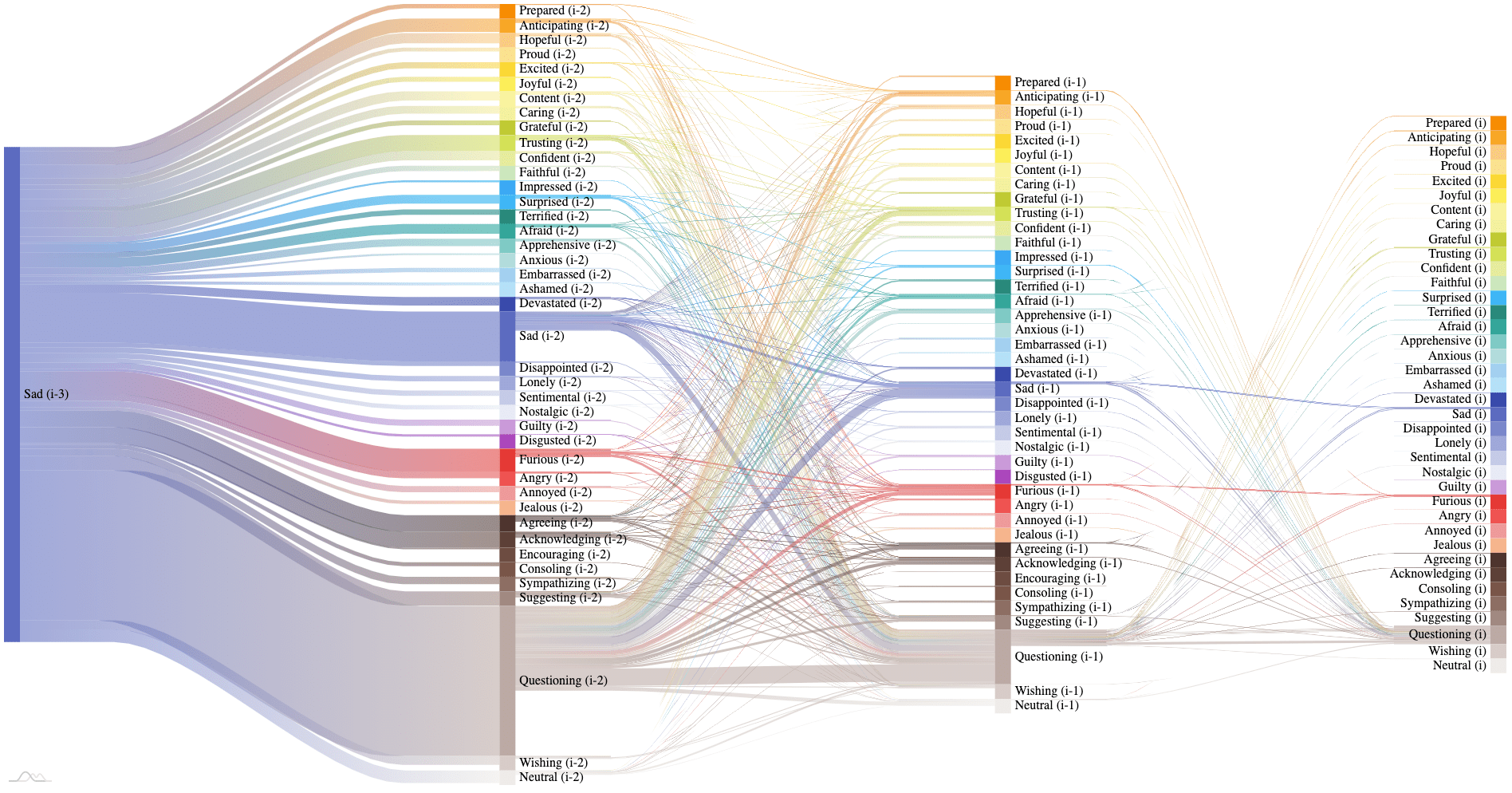} }}%
    \qquad
    \subfloat[\centering Beginning emotion: Angry]{{\includegraphics[width=0.45\textwidth]{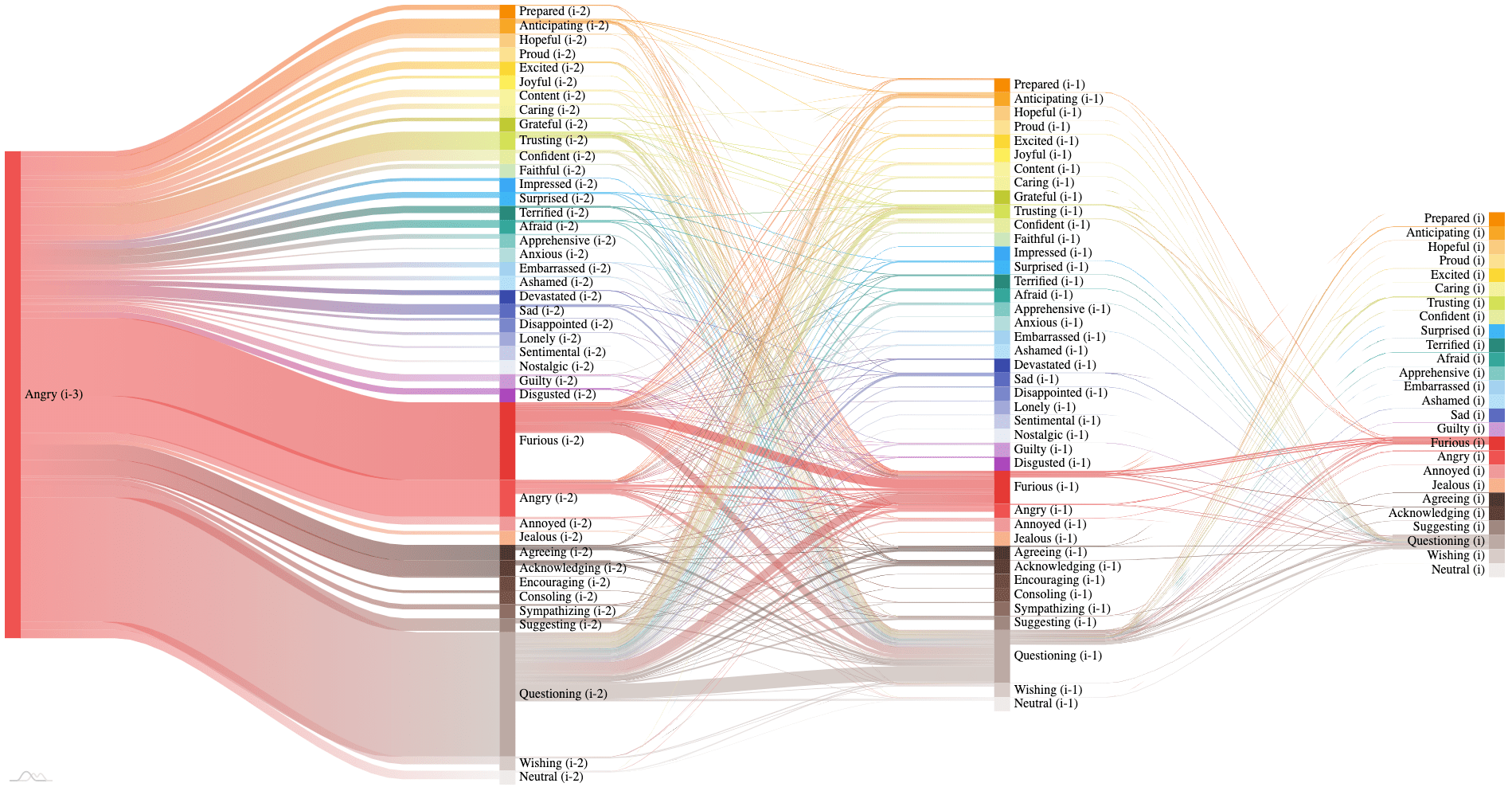} }}%
    
    \caption{Emotion-intent flow visualizations in the OSED dataset.}%
    \label{fig:visualizations}%
\end{figure*}

OSED with 1M automatically annotated dialogues could be readily used in applications such as developing chatbots that could generate controlled responses conditioned on emotions and intents. This requires an evaluation of the annotation quality of the resultant dataset. Though the accuracy of the EB+ classifier speaks for OSED annotation quality, we performed a visual analysis on the emotion-intent flow patterns in OSED to inspect if they agree with human social interaction patterns.

 
 Figure \ref{fig:visualizations} shows emotion-intent flow patterns that occur in OSED when the beginning dialogue emotion is one of \textit{joyful}, \textit{surprised}, \textit{sad} and \textit{angry}, respectively. We limit the patterns to 4 dialogue turns for ease of visualization. Visualizations for further emotions are included in Appendix \ref{sec:app_visualizations}. According to the visualizations, the most frequent intent that immediately follows a given emotion, irrespective of the emotion being positive or negative, is \textit{questioning}. This observation is on par with claims made by See et al. \shortcite{see} and Welivita and Pu \shortcite{taxonomy}. Intents \textit{acknowledging} and \textit{agreeing} also take prominence in turns that follow emotional dialogue turns. Another important observation is that emotional dialogue turns are frequently followed by turns containing similar emotions. For example, \textit{joyful} followed by emotions \textit{excited}, \textit{joyful}, and \textit{content}, which belong to the same sector in the Plutchik's emotion wheel \cite{plutchik}. This is a bit contradictory to observations made by Welivita and Pu \shortcite{taxonomy} in EmpatheticDialogues, in which they state the listener responses are mostly neutral and contain specific intents such as \textit{questioning}, \textit{agreeing} or \textit{acknowledging}. However, our observation here can be justified since, in EmpatheticDialogues, the listeners were specifically asked to react calmly and reassuringly, whereas movie dialogues are more natural and dramatic.



\section{Discussion}

This study proposed an emotional dialogue curation pipeline from movie subtitles resulting in 1M emotional movie dialogues automatically annotated with 32 emotions and 9 empathetic response intents. The new response intents such as \textit{Greeting}, \textit{Declaration}, \textit{Commanding}, \textit{Hesistant}, and \textit{Accusing} proposed by the crowd workers, validate the complexity of listeners' reactions to emotions and invoke the need to further conduct analysis in this area. The fine-grained dialogue emotion classifier we utilized in the annotation process could be used as a general-purpose classifier for human social conversations. An F1-score of $64\%$ over $41$ class labels is also significant. Even the top-performing systems in recent EmoContext, EmotionX 2018, and EmotionX 2019 challenges, which also use pre-trained BERT in their model architecture, report F1-scores in the range $88.5\%$ (EmotionX-IDEA \cite{emotionx-idea} on EmotionPush dataset)  - $76\%$ (FigureEight \cite{figure-8} on EmoContext dataset). Since these scores are reported only on a limited number of emotions, the F1-score of $64\%$ we report in this paper over $41$ emotion and intent classes is comparable to the performance of the state-of-the-art. The visualization of emotion-intent exchange patterns further suggests the resultant OSED dataset conforms to the emotion-intent exchange patterns in human social conversations and can be utilized in applications such as developing chatbots that could learn and respond to emotional prompts. 

Implications of this work include 1) help in assessing the response quality of existing social chatbots at a finer granularity, 2) inform social chatbots about the desirable responses given an emotional prompt, and 3) help in the design and development of more controllable and interpretable neural chatbots \cite{Towards_explainable_and_controllable_open_domain_dialogue_generation_with_dialogue_acts}. There are some limitations to this approach as well, which include inaccuracies occurring due to automatic turn and dialogue segmentation and annotation with only 9 empathetic response intents, whereas in reality, there can be more and even contrasting intents such as \textit{Accusation}, and \textit{Disagreement}. 

\section{Conclusion}

In this work, we curated a large-scale dialogue dataset, OSED, comprising of 1M emotional dialogues from movie subtitles. This dataset is more general-purpose, larger in size, and contains more fine-grained emotion categories and empathetic response intents than the existing emotional dialogue datasets. To facilitate annotation, we developed a dialogue emotion classifier capable of recognizing 32 fine-grained emotions and 9 empathetic response intents with significant accuracy. It was trained on movie dialogues initially annotated using human computation and extended using self-labeling, and sentence similarity approaches. As future work, we intend to extend the taxonomy of empathetic response intents using new labels discovered during this process and utilize the OSED dataset to develop a controllable neural chatbot capable of generating empathetic responses during social chitchat. 

\begin{quote}
\small
\bibliographystyle{acl_natbib}
\bibliography{acl2020}
\end{quote}

\appendix

\section{Computing the readability of OS dialogues}
\label{sec:app_readability}

We followed the following steps in calculating the readability of the dialogues. The dialogues that scored high in readability were preferred for the crowd-annotation task since they avoid the overhead of having to read long and complex dialogues that may exhaust the crowd-worker.

\begin{enumerate}
    \item Build a frequency vocabulary by calculating the token count for all the dialogues in the cleaned OS dataset.
    \item For each dialog, aggregate the frequencies of all tokens and take the average using the following formula, in which $f_{sum}$ is the sum of frequencies of all tokens, $n_{tokens}$ is the total number of tokens in the dialog, and $\alpha$ is a constant (set to 87 in our case). The idea behind this is that difficult to read dialogues contain less frequent words and should result in less readability.   
    \[ f = f_{sum} / (\alpha + n_{tokens}) \]
    \item For each dialog, also calculate the percentage of distinct words, say $d$. 
    \item Finally, compute the readability score for each dialogue by taking the weighted sum of $f$ and $d$. Experimental results showed that the combination of $f + 0.04d$ was giving the best results. We take the combination of both $f$ and $d$ because, if only $f$ is considered, then dialogues that contain a lot of repetitive tokens can score high in readability, which is undesirable. 
\end{enumerate}

\section{AMT task interfaces}
\label{sec:app_interface}

\begin{figure}[ht!]
\centering
\includegraphics[width=\linewidth]{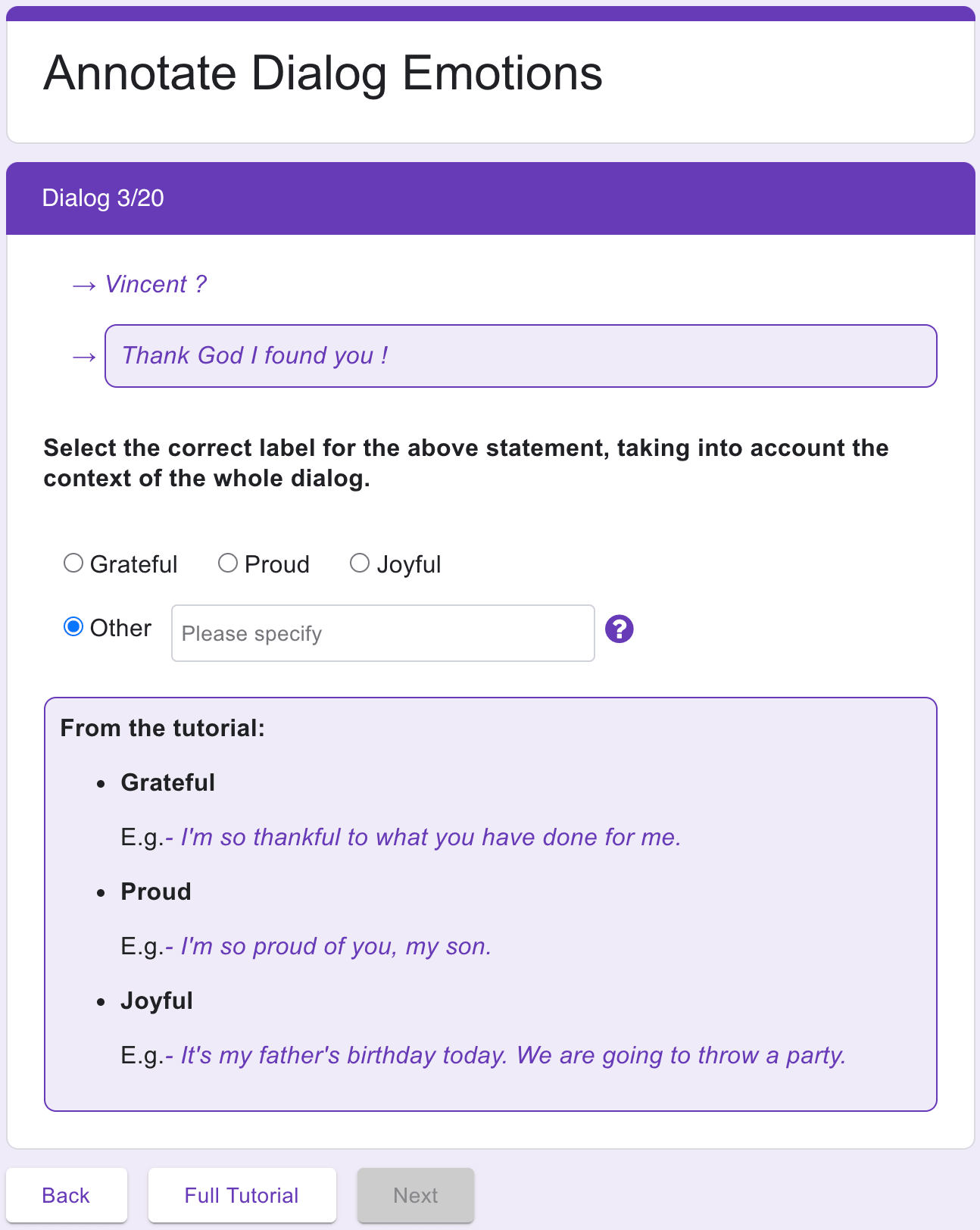} 
\caption{The user interface of the AMT crowd-annotation task.}
\label{fig:mturk}
\end{figure}

\begin{figure}[ht!]
\centering
\includegraphics[width=\linewidth]{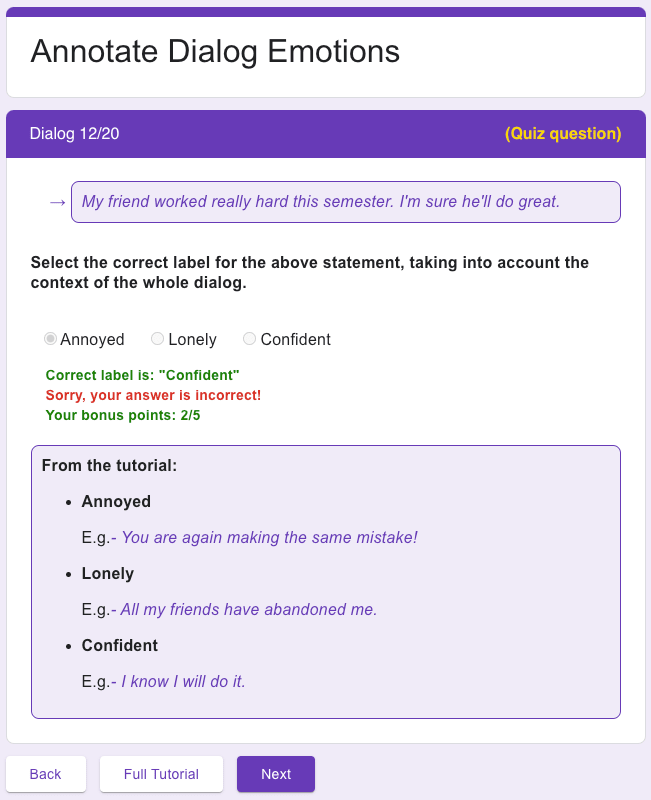} 
\caption{Quiz question interface design of the AMT crowd-annotation task. Quiz questions were used to measure the quality of the crowdworkers as well as help them improve their annotation quality by understanding their mistakes.}
\label{fig:mturk_quiz}
\end{figure}

The user interface used to collect labels from the AMT workers is denoted in Figure \ref{fig:mturk} and the design of quiz question interface is denoted in Figure \ref{fig:mturk_quiz}. 

\section{New intent categories discovered in the AMT task}
\label{sec:app_newlabels}

Table \ref{table:new_categories} lists some of the frequent classes proposed by the crowd workers during the AMT crowd-annotation task along with corresponding examples.  

\begin{table}[ht!]
\centering
\begin{tabularx}{\linewidth}{|l|X|}
\hline
New category & Example dialogue \\
\hline

Greeting &
{\begin{tabularx}{\linewidth}{X}
\textit{- Hello, Florelle.}
\end{tabularx}}\\

\hline

Declaration &
{\begin{tabularx}{\linewidth}{X}
\textit{- Claire, what 's happening?}\\
\textit{- I saw him. He was here.}
\end{tabularx}} \\

\hline

Commanding &
{\begin{tabularx}{\linewidth}{X}
\textit{- Power of unity!}\\
\textit{- Alpha Team, it's a combined operation. After putting on winter military gear, come to 464 airport by 21:00 today.}\\
\end{tabularx}} \\

\hline

Hesistant &
{\begin{tabularx}{\linewidth}{X}
\textit{- Now that's more like it.}\\
\textit{- Where have you been all night?}\\
\textit{- I just ... went out for a walk.}
\end{tabularx}} \\

\hline

Accusing &
{\begin{tabularx}{\linewidth}{X}
\textit{- I don't want anything to do with your fucked-up, exploitative, supplying God knows how many black markets school. OK?} \\
\textit{-You should be ashamed, making money out of a captive audience.} \\
\end{tabularx}} \\

\hline

\end{tabularx}
\caption{New categories proposed by the crowd workers along with corresponding examples.}
\label{table:new_categories}
\end{table}

\section{Choice of hyper-parameters and additional training details regarding EB+}
\label{sec:app_params}

\begin{figure}[ht!]
\centering
\includegraphics[width=\linewidth]{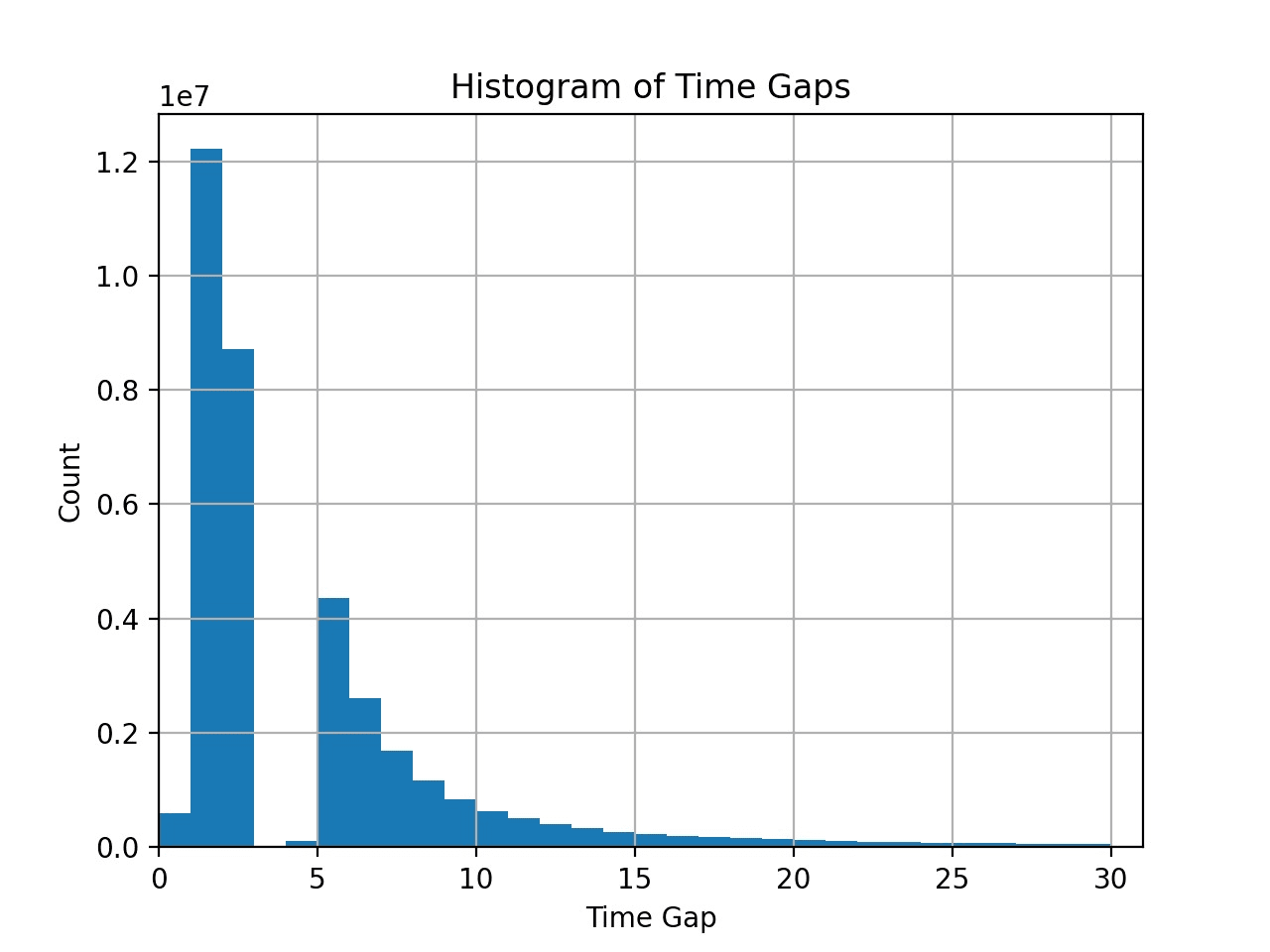} 
\caption{Histogram of time intervals between adjacent subtitle blocks in the OpenSubtitles corpus.}
\label{fig:histogram}
\end{figure}

The choice of 5 seconds to separate dialogues is based on a histogram of time intervals between adjacent subtitle blocks in the OpenSubtitles corpus, which is denoted in Figure \ref{fig:histogram}. As it can be observed in the histogram, most of the time gaps fall below 3 seconds. A clear drop in count was observed between 3-5 seconds. Therefore, we chose 5 seconds as the time interval to separate dialogues.   

Using decreasing weights for context utterances is based on the intuition that in human dialogues, more attention is paid to the most recent utterances in dialogue history. This idea is backed up by time-decay functions used in neural dialogue understanding approaches \cite{see}. We conducted an ablation study with and without using decreasing weights in the model. Performance of the unweighted models was lower than the performance of weighted models yielding final F1 scores of $63.44$ and $64.86$ for unweighted and weighted models, respectively. 

All the experiments were conducted on a machine with 2x12cores@2.5GHz, 256 GB RAM, 2x240 GB SSD, and 2xGPU (NVIDIA Titan X Maxwell).  

\section{Distribution of labels in the datasets used to train the EB+ classifier}
\label{sec:app_distribution}

\begin{figure}[ht!]
\centering
\includegraphics[width=\linewidth]{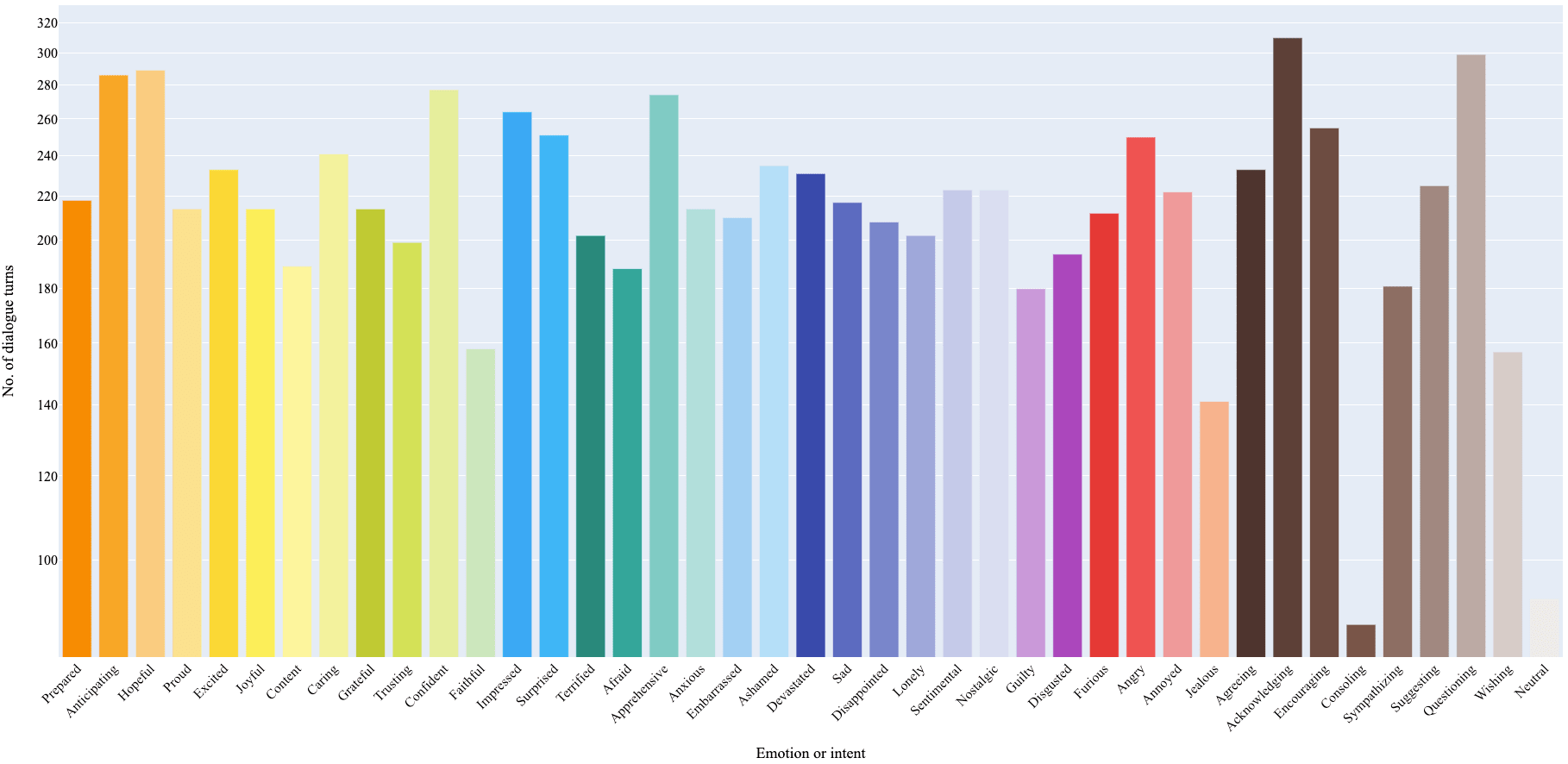} 
\caption{Distribution of emotions and intents in the AMT annotated 9K OS dialogues.}
\label{fig:dist_1}
\end{figure}

\begin{figure}[ht!]
\centering
\includegraphics[width=\linewidth]{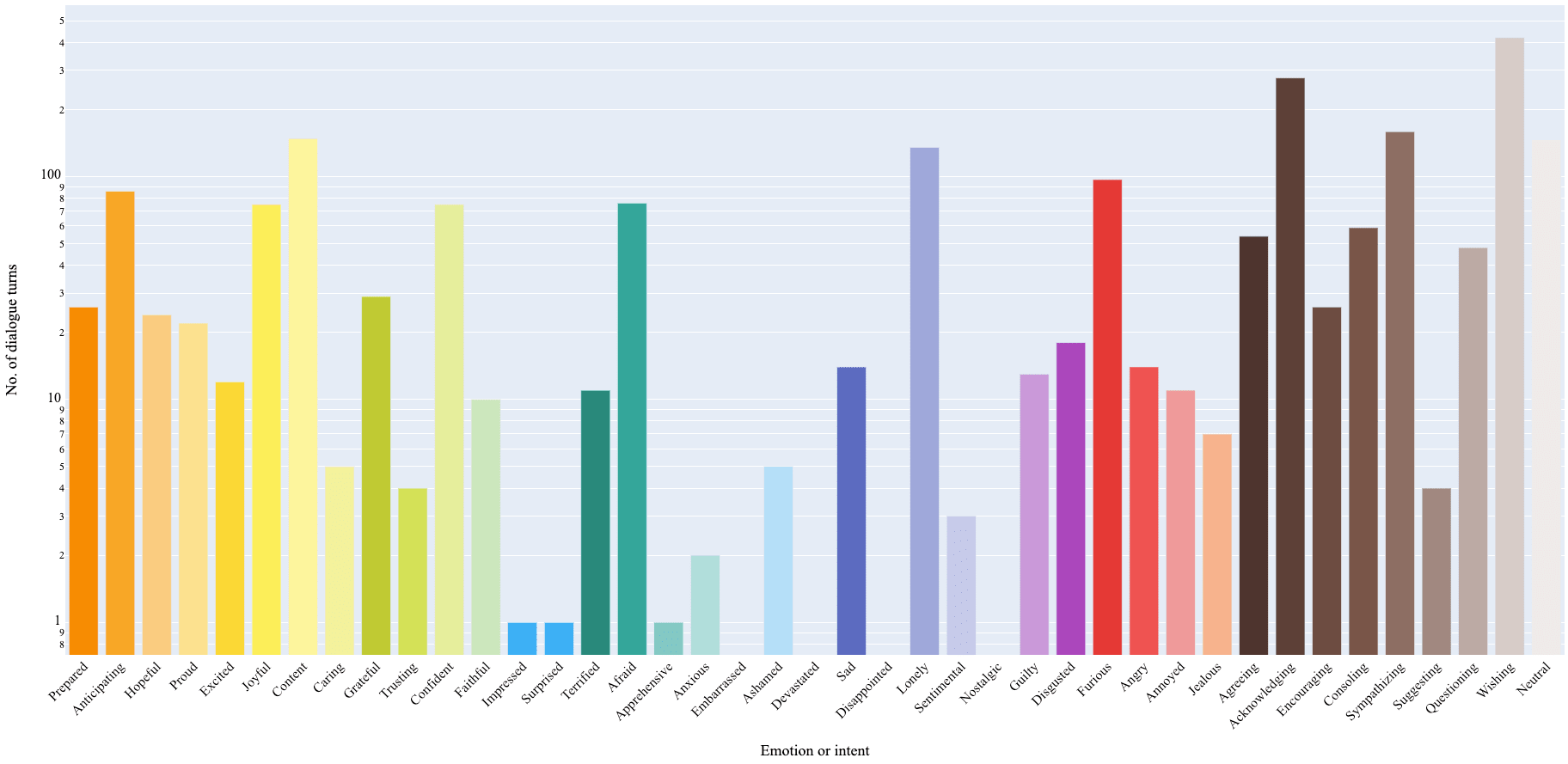} 
\caption{Distribution of emotions and intents in the 2K similar self-labeled dialogues incorporated in the EB+ 2nd iteration, extended.}
\label{fig:dist_2}
\end{figure}

Figure \ref{fig:dist_1} shows the distribution of emotions and intents in the crowd-annotated 9K dialogues used to train the EB+ classifier, while Figure \ref{fig:dist_2} shows the distribution of labels extended by considering dialogue similarity using the Siamese-BERT approach \cite{siamese}. 

\section{Comparison of the performance of the EB+ classifier with the state-of-the-art dialogue emotion classifiers}
\label{sec:app_performance_comparison}

Table \ref{table:perf_com} compared the performance of the EB+ classifier with several state-of-the-art dialogue emotion classifiers. However, a direct comparison cannot be made since existing classifiers are trained to predict only a limited set of emotion labels. 

\begin{table*}[ht!]
\centering
\begin{tabularx}{\textwidth}{|X|X|X|r|r|r|r|}
\hline
Classifier & Dataset & No. of labels & F1 & Acc.\\
\hline

AR \cite{emotionx-ar} & EmotionLines  & 4 Emotion labels & $-$ & Friends: $62.50$\\
 & dataset \cite{emotionlines} &  & & EmotionPush: $62.48$\\
\hline
CMN \cite{cmn} & IEMOCAP dataset \cite{iemocap} & 6 Emotion labels & $56.13$ & $56.56$\\
\cline{1-1} \cline{4-5}
ICON \cite{icon} & & & $57.90$ & $58.30$\\
\cline{1-1} \cline{4-5}
IAAN \cite{iaan} & & & $-$ & $64.70$\\
\hline
Dialog-RNN \cite{dialoguernn} & IEMOCAP \cite{iemocap} and AVEC \cite{avec} datasets & IEMOCAP: 4 Emotion labels; AVEC: 4 dimentional emotion labels & $62.75$ & $63.40$\\
\hline
Dialog-GCN \cite{dialoguegcn} & IEMOCAP \cite{iemocap}, AVEC \cite{avec}, and MELD \cite{meld} datasets & IEMOCAP: 4 Emotion labels; AVEC: 4 dimentional emotion labels; MELD: 7 Emotion labels & $64.18$ & $65.25$\\
\hline
ANA \cite{ana} & EmoContext dataset \cite{emocontextdataset} & 4 Emotion labels & $82.86$ & $-$\\
\hline
EB+ (2\textsuperscript{nd} iter., extended) & OS dialogue dataset & 32 Emotion labels + 9 Intent labels & $63.86$ & $65.00$\\

\hline
\end{tabularx}
\caption{Comparison of the performance results of the EB+ classifier (2nd iter., extended) with performance of the state-of-the-art dialogue emotion classifiers. F1-score reported here is the macro-F1 score.}
\label{table:perf_com}
\end{table*}


\section{Distribution of emotions and intents in the OSED dataset}
\label{sec:app_distribution_osed}

\begin{figure}[ht!]
\centering
\includegraphics[width=\linewidth]{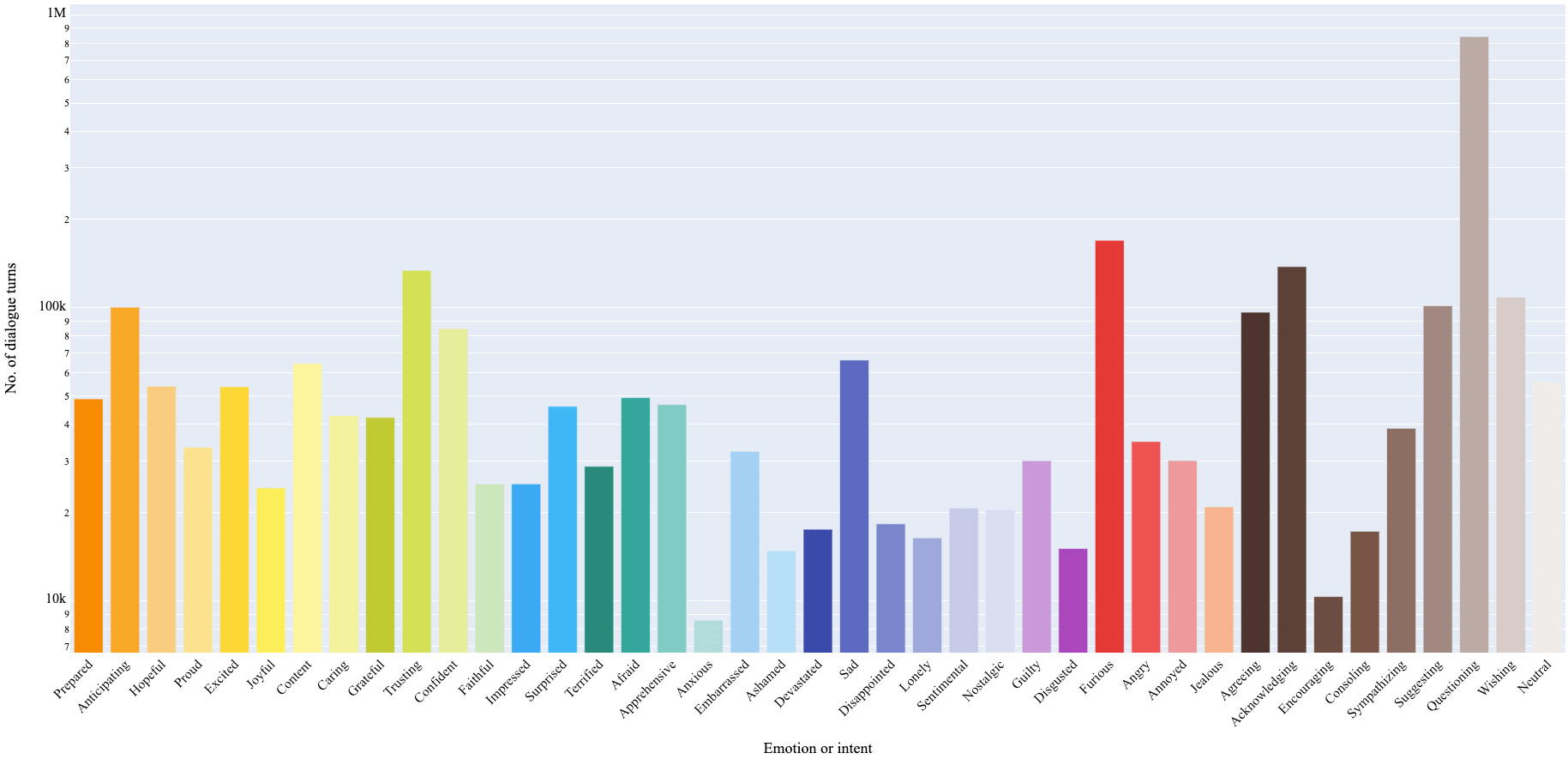} 
\caption{Distribution of emotions and intents in the OSED dataset. The numbers are in the log scale.}
\label{fig:osed_dist}
\end{figure}

Figure \ref{fig:osed_dist} shows the distribution of emotion and intent labels in the resultant 1M OSED dataset. It could be noted that intent categories take prominence over individual emotion classes, which is expected since one or more intents could be utilized when responding to emotions in dialogue. 

\section{Example dialogues from OSED along with EB+ labels and confidence scores.}
\label{sec:app_examples}

Table \ref{table:osed_example} denotes some example dialogues taken from the OSED dataset along with EB+ labels predicted for each dialogue turn and their confidence values.


\begin{table*}[ht!]
\centering
\begin{tabularx}{\textwidth}{|X|}
\hline


- \textit{I have to catch a flight} (Anticipating, 0.85)\\
- \textit{Okay , Robert} (Acknowledging, 0.73) \\

\hline

- \textit{Completely empty .} (Lonely, 0.73) \\
- \textit{Cheers . How hungry I was . Dig in .} (Encouraging, 0.78) \\
\hline



- \textit{Thank you for coming , Lleó . Thanks for coming to see me on your birthday.} (Grateful,	0.77) \\
- \textit{I appreciate that . Gavina has broken her left foot and you 've come to visit her .}  (Grateful, 0.97) \\

\hline

- \textit{I 've been wanting to tell you something . I 'd admire you so much The stand you 're taking . You didn 't take the easy way out .} (Impressed, 0.88) \\
- \textit{Not setting up one of your friends . I couldn 't do what you did . So looks like you 're the one .. .. teaching me . What real character and integrity is all about . I love you son .} (Proud, 0.43) \\
\hline

- \textit{Could you please hurry up ?} (Suggesting, 0.69) \\ 
- \textit{I 'm doing my best , miss . I have a feeling Elena will be here any minute .} (Hopeful, 0.48) \\
- \textit{Let her !} (Encouraging,	0.51)
- \textit{We 're not doing anything wrong.} (Content, 0.47) \\
- \textit{Done ! Wonderful !} (Acknowledging, 0.69) \\
- \textit{Here is your money . My sister will show you to the door .} (Trusting, 0.62) \\
\hline

\end{tabularx}
\caption{Example dialogues from the OSED dataset along with EB+ annotations and confidence scores.}
\label{table:osed_example}
\end{table*}


\section{Visualizations of emotion-intent flow patterns in the OSED dataset}
\label{sec:app_visualizations}

\begin{figure}[h!]
\centering
\includegraphics[width=\linewidth]{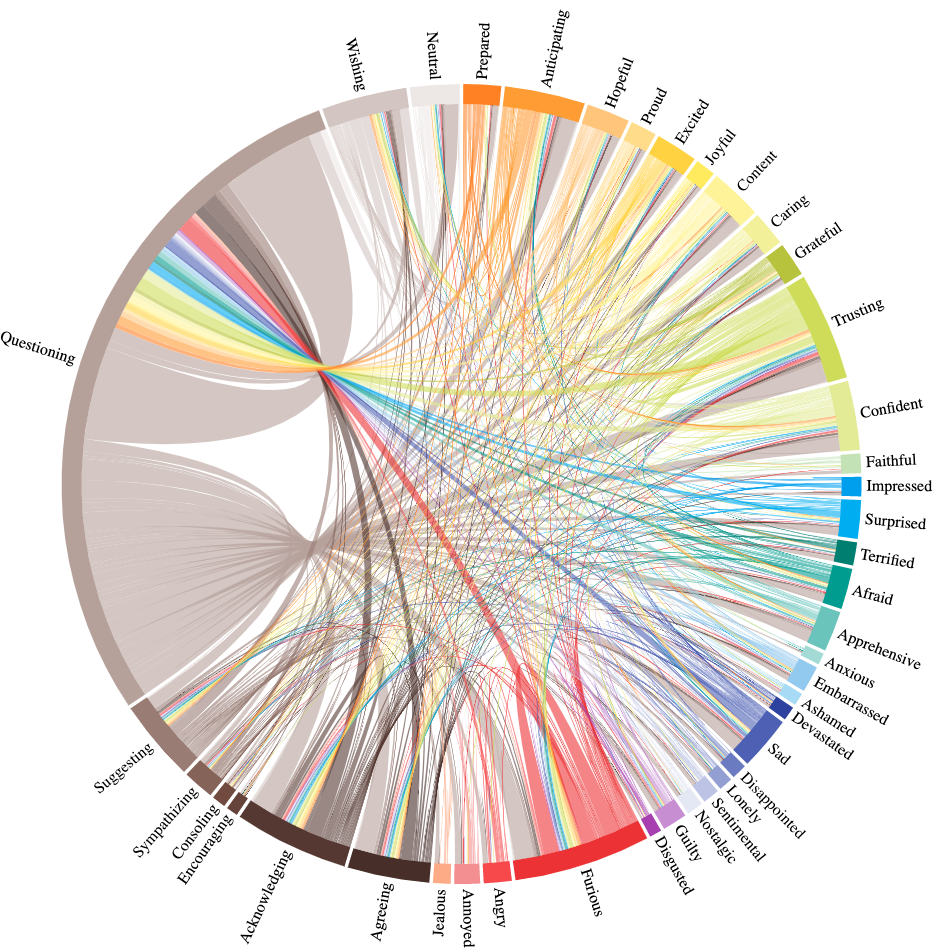} 
\caption{Emotion-intent exchanges between adjacent dialogue turns in the OSED dataset.}
\label{fig:simple_visu}
\end{figure}

\begin{figure*}[h!]%
    \centering
    
    \subfloat[\centering Beginning emotion: Anticipating]{{\includegraphics[width=0.45\textwidth]{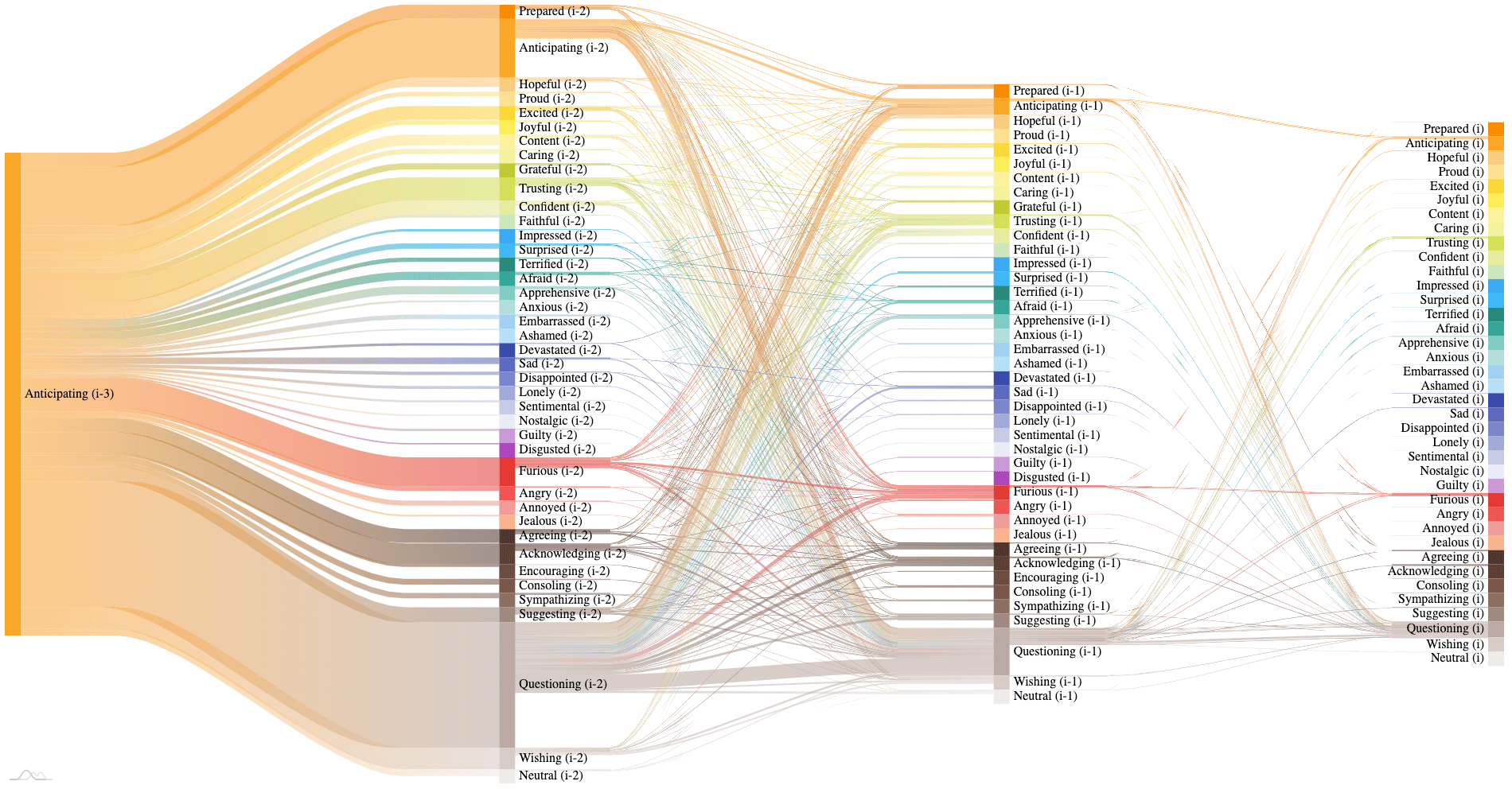} }}%
    \qquad
    \subfloat[\centering Beginning emotion: Trusting]{{\includegraphics[width=0.45\textwidth]{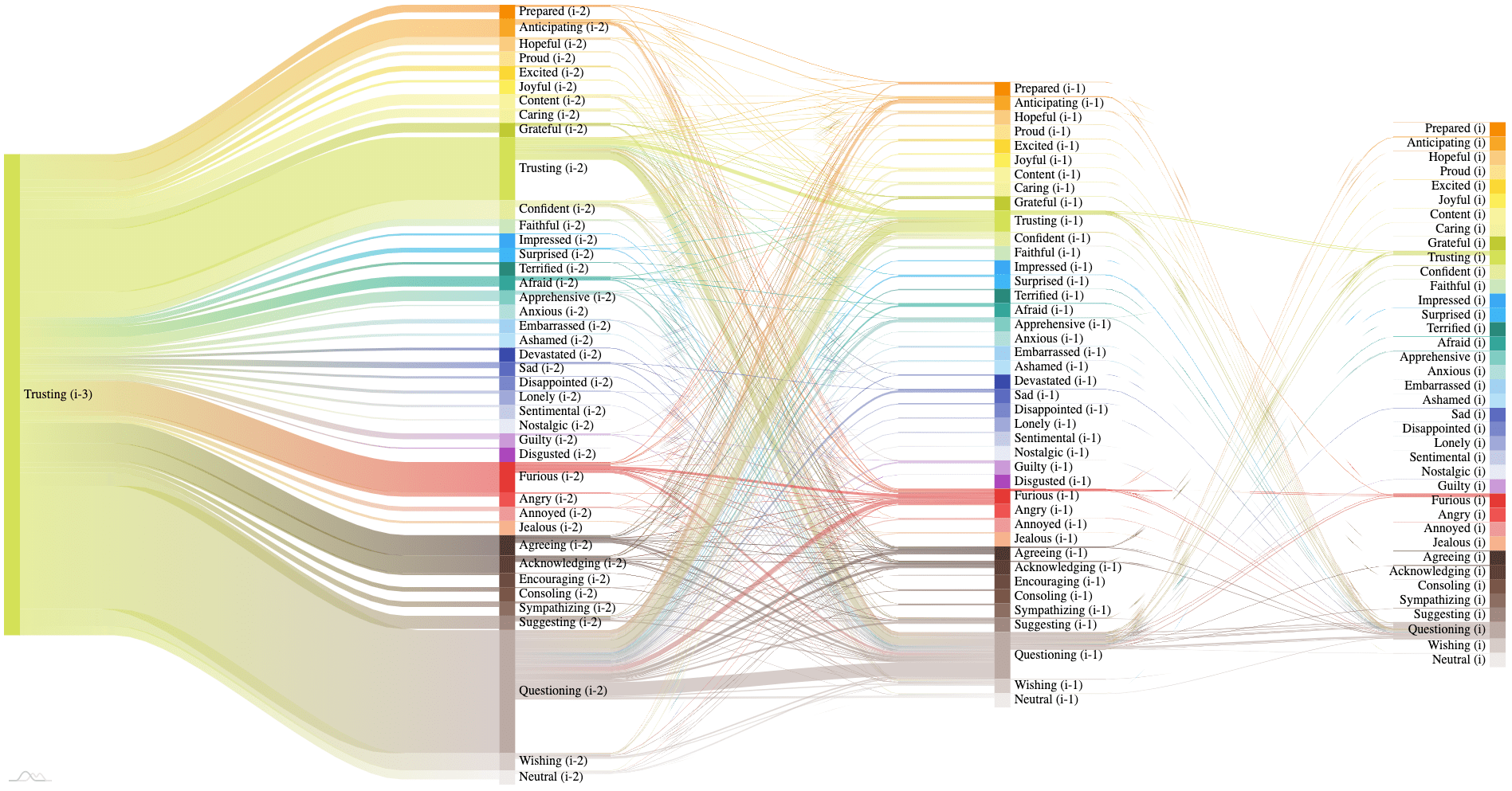} }}%
    
    \subfloat[\centering Beginning emotion: Afraid]{{\includegraphics[width=0.45\textwidth]{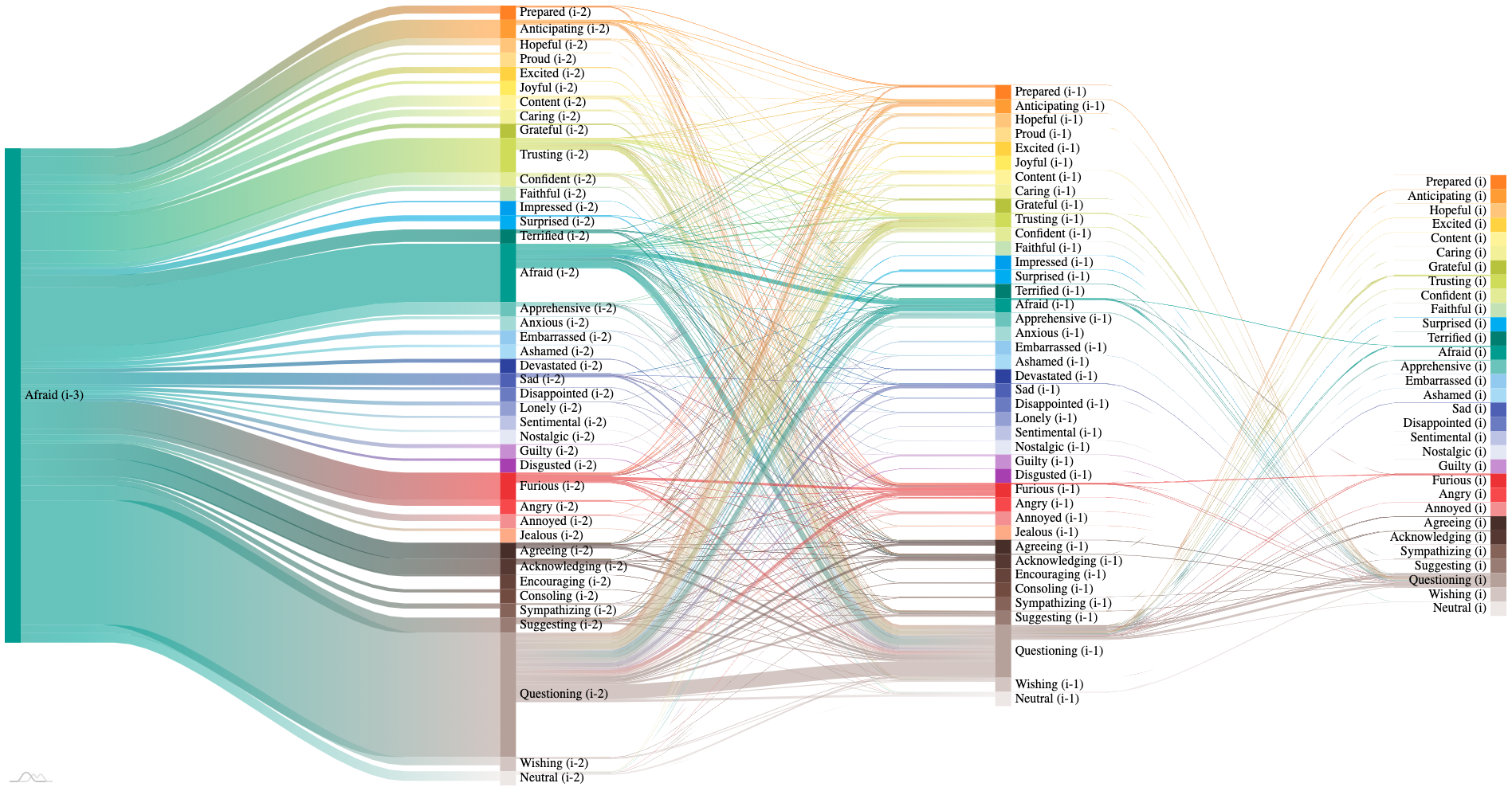} }}%
    \qquad
    \subfloat[\centering Beginning emotion: Disgusted]{{\includegraphics[width=0.45\textwidth]{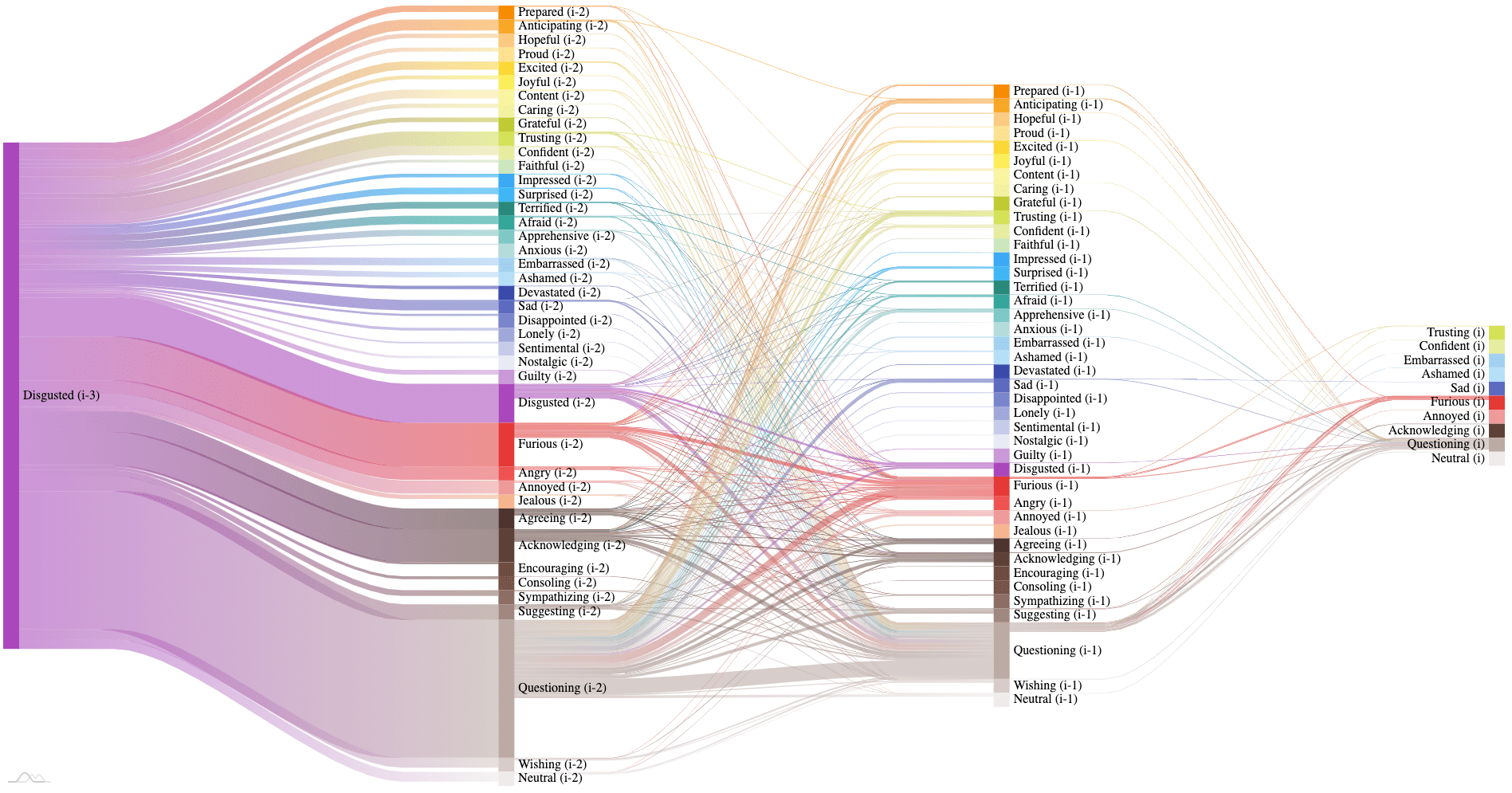} }}%
    
    \caption{Emotion-intent flow visualizations in the OSED dataset when the beginning dialogue emotion is one of \textit{anticipating}, \textit{trusting}, \textit{afraid} or \textit{disgusted}.}%
    \label{fig:visualizations_2}%
\end{figure*}

Figure \ref{fig:simple_visu} shows emotion-intent exchanges between adjacent dialogue turns in the OSED dataset. Figure \ref{fig:visualizations_2} denotes some additional emotion-intent flow patterns in the OSED dataset, when the beginning dialogue emotion is one of \textit{anticipating}, \textit{trusting}, \textit{afraid} or \textit{disgusted}.  They could be used to further justify the observations made in Section \ref{sec:quality}.

\end{document}